\documentclass[10pt,twocolumn,letterpaper]{article}

\usepackage{cvpr} 
\usepackage{times}
\usepackage{epsfig}
\usepackage{graphicx}
\usepackage{amsmath}
\usepackage{amssymb}
\usepackage{multirow}
\usepackage{enumitem}
\setitemize{noitemsep,topsep=0pt,parsep=0pt,partopsep=0pt}
\usepackage{subcaption}
\usepackage{makecell}
\usepackage{bm}
\usepackage{booktabs}

\usepackage{xcolor}
\usepackage{hyperref}
\definecolor{modernCite}{RGB}{0, 160, 140}
\definecolor{modernLink}{RGB}{140, 0, 210}  
\definecolor{modernURL}{RGB}{220, 100, 0}   

\hypersetup{
    colorlinks=true,         
    citecolor=modernCite,   
    linkcolor=modernLink,   
    urlcolor=modernURL,   
    pdftitle={CanonicalPhys: Pose-Robust Remote Photoplethysmography via Canonical-Space Priors},
}

\begin{document}

\title{CanonicalPhys: Pose-Robust Remote Photoplethysmography via Canonical-Space Priors}

\author{Hui Wei$^{1,2}${\thanks{Equal contribution. $^{\dagger}$Corresponding author}} \,\, Seyedata Jodeiri Seyedian$^{1,2}$\footnotemark[1] \,\, Xiaobai Li$^{3}$ \,\, Guoying Zhao$^{2,1\dagger}$\\
\small $^{1}$Center for Machine Vision and Signal Analysis (CMVS), University of Oulu\\
\small $^{2}$ELLIS Institute Finland\\
\small $^{3}$Zhejiang University\\
{\tt\small \{hui.wei, ata.seyedian, guoying.zhao\}@oulu.fi, xiaobai.li@zju.edu.cn}
}

\maketitle
\thispagestyle{empty}

\begin{abstract}
Deep remote photoplethysmography (rPPG) attains sub-bpm heart-rate error on frontal, stationary faces yet degrades sharply under head pose: on MMPD, the state-of-the-art FactorizePhys backbone's MAE grows $1.60\times$ from frontal ($|\text{yaw}|{<}15^\circ$) to large-yaw ($|\text{yaw}|{\geq}45^\circ$) frames. We argue that pose is a \emph{coordinate-structural} nuisance rather than a data-augmentation problem: in image coordinates the same pixel maps to different anatomy at different poses, blocking three priors otherwise natural for rPPG, namely the dichromatic reflection model, pulse-phase invariance across skin regions, and the POS/CHROM chromaticity projection, each of which presumes a stable anatomy-to-pixel mapping. We introduce \textbf{CanonicalPhys}, which prepends a differentiable four-point homography that fixes four facial anchors at canonical positions; in this canonical frame the three priors become expressible as a per-pixel Lambertian weight, a cross-ROI temporal consistency loss, and knowledge distillation from windowed POS, none of which adds trainable parameters over the backbone. At an identical parameter count, CanonicalPhys reduces MMPD's frontal-to-large-yaw MAE degradation from $1.60\times$ to $1.33\times$ and flattens the mild-yaw bin from $1.32\times$ to $1.07\times$ (across CanonicalPhys variants), with matched cross-dataset MAE reductions of up to $32\%$ on pose-rich targets. Code: \url{https://github.com/infraface/CanonicalPhys}.
\end{abstract}

\section{Introduction}

\begin{figure}[t]
    \centering
    \includegraphics[width=\columnwidth]{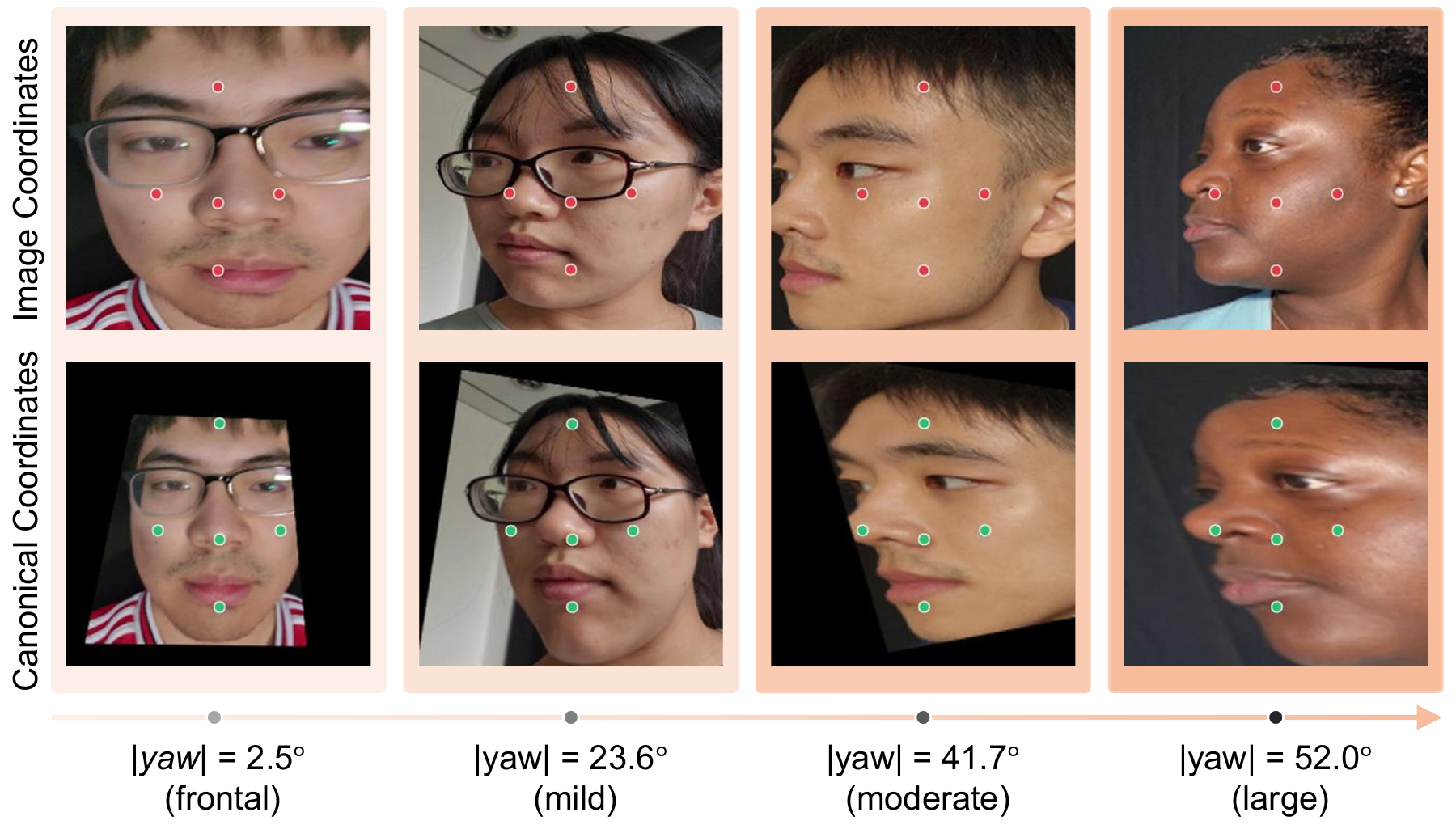}
    \vspace{-5mm}
    \caption{\textbf{Pose is a coordinate-structural nuisance for rPPG.} (Top) On MMPD, as head yaw grows, a fixed image-coordinate pixel drifts across forehead, eyebrow, and temple; any architecture that reasons in image coordinates cannot assign a stable anatomical role to any pixel. (Bottom) Our proposed CanonicalPhys prepends a four-point homography that maps four facial anchors to fixed canonical positions, approximately restoring the dominant rigid component of the anatomy-to-pixel mapping.}
    \label{fig:teaser}
    \vspace{-5mm}
\end{figure}

Remote photoplethysmography (rPPG) estimates the blood volume pulse from ordinary facial video, enabling contact-free cardiovascular sensing in telemedicine, neonatal care, and driver-state monitoring~\cite{verkruysse2008remote,chen2018deepphys,Liu_2023_WACV}. Modern deep rPPG models~\cite{yu2019remote,Liu_2023_WACV,yu2022physformer,joshi2024factorizephys} attain sub-bpm (beats per minute) MAE on frontal, stationary video, yet they degrade sharply when the subject turns the head. On MMPD, the FactorizePhys backbone's MAE grows $1.60\times$ from frontal ($|\text{yaw}|{<}15^\circ$) to large-yaw ($|\text{yaw}|{\geq}45^\circ$) frames (Sec.~\ref{sec:pose}). Because realistic deployments routinely contain non-frontal frames, pose-induced degradation directly bounds the practical utility of deployed rPPG. Figure~\ref{fig:teaser} visualizes this gap: as yaw grows, image-coordinate pixels drift across anatomy, and the baseline's MAE rises monotonically, while CanonicalPhys stays flat.

We argue that this is a \emph{coordinate-structural} problem rather than a data-augmentation one. An image-coordinate pixel $(x,y)$ maps to different anatomy at different poses, which blocks three priors that are otherwise natural for rPPG. First, the dichromatic reflection model~\cite{wang2016algorithmic} predicts that the diffuse, pulse-carrying component of skin reflectance scales as the cosine between the skin normal and the camera axis, a quantity defined per \emph{anatomical} location rather than per pixel. Second, pulse-phase invariance across forehead and cheeks requires a pose-stable identification of each ROI. Third, classical POS and CHROM~\cite{wang2016algorithmic,de2013robust} project windowed RGB traces onto a chromaticity basis, a projection well-posed only on a fixed skin ROI. A stable anatomy-to-pixel mapping makes all three tractable again.

Motivated by this observation, \textbf{CanonicalPhys} prepends a differentiable four-point homography on the outer eye and mouth corners that maps each frame into a canonical face coordinate system.\footnote{Throughout, \emph{no added parameters} (or \emph{parameter-free}) means that no \emph{trainable} parameters are added over the FactorizePhys backbone. Canonicalization and the Lambertian weight consume 2D landmarks and a facial-transform matrix from a frozen, off-the-shelf MediaPipe~\cite{lugaresi2019mediapipe} model, precomputed once per video, on par with the RetinaFace~\cite{deng2020retinaface} face detector already required by every baseline for cropping.} On top of an unmodified FactorizePhys backbone, we attach three canonical-space components: a per-pixel Lambertian weight from canonical normals and the per-frame MediaPipe~\cite{lugaresi2019mediapipe} rotation; a cross-ROI temporal consistency loss over fixed canonical forehead and cheek features; and knowledge distillation from POS applied to canonical forehead pixels. None adds trainable parameters, so the parameter count is identical to the backbone.

Pose-stratified evaluation on MMPD (Sec.~\ref{sec:pose}) reduces the frontal-to-large-yaw MAE degradation from $1.60\times$ to $1.33\times$ and flattens the mild-yaw bin from $1.32\times$ to $1.07\times$. Cross-dataset evaluation on UBFC-rPPG~\cite{bobbia2019unsupervised}, PURE~\cite{stricker2014non}, OBF~\cite{li2018obf}, and MMPD~\cite{tang2023mmpd} confirms that the effect transfers across sensors. On the pose-richer UBFC-trained targets, the matched-seed MAE reductions are $32\%$ on PURE and $20\%$ on MMPD, PURE seed variance is more than halved, and CanonicalPhys wins $13$ of the $16$ cells of the full transfer matrix.
In summary, our contributions are:
\begin{enumerate}[leftmargin=*]
    \item We introduce a differentiable four-point homography, with no trainable parameters, that maps each frame into a canonical face coordinate system, absorbing the dominant rigid component of head pose before any learned processing and approximately restoring a pose-stable anatomy-to-pixel mapping.
    \vspace{-2mm}
    \item We propose three canonical-space components enabled by this change of coordinates, none of which adds trainable parameters: a physics-informed per-pixel Lambertian weight from the dichromatic reflection model, a cross-ROI temporal consistency loss that enforces pulse-phase invariance, and a knowledge distillation loss from windowed POS applied to canonical forehead pixels.
    \vspace{-2mm}
    \item We evaluate CanonicalPhys with a matched-seed pose-stratified analysis on MMPD that directly isolates pose from other distribution shifts, a full cross-dataset transfer across UBFC-rPPG, PURE, OBF, and MMPD reporting the complete metric suite, and a failure-mode analysis (Sec.~\ref{sec:failure_modes}) that delineates the five regimes where the claim does not yet hold: moderate yaw, large yaw, extreme yaw, landmark fallback, and pitch/roll.
\end{enumerate}

\section{Related Work}

\noindent\textbf{Deep rPPG.}
Early deep rPPG methods adapted convolutional architectures from generic video understanding to blood volume pulse regression. PhysNet~\cite{yu2019remote} introduced a 3D-CNN encoder-decoder trained with a temporal Pearson correlation loss. DeepPhys~\cite{chen2018deepphys} and TS-CAN~\cite{liu2020multi} proposed a two-stream appearance-motion design. EfficientPhys~\cite{Liu_2023_WACV} replaced 3D convolutions with temporal shift modules and a self-attention shifted network head. Transformer-based PhysFormer~\cite{yu2022physformer} introduced temporal-difference attention. More recently, FactorizePhys~\cite{joshi2024factorizephys} computes joint spatial-temporal-channel attention via nonnegative matrix factorization and remains state-of-the-art among lightweight models. All of these methods learn their spatial priors purely from raw appearance, and none consult an external geometric signal. As a consequence, a pixel at coordinate $(x,y)$ plays a different anatomical role at different head poses, and the pose nuisance is left to the backbone to absorb through capacity alone.

\noindent\textbf{Cross-dataset generalization for rPPG.}
The dominant strategy for closing the cross-dataset gap is distributional alignment in a learned feature space. FreqPhys~\cite{qian2026freqphys} repurposes an implicit physiological frequency prior as a regularizer during training. HOT~\cite{nguyen2026hot} formulates cross-dataset adaptation as harmonic-constrained optimal transport between source and target representations. 
We take an orthogonal route and inject a prior that is domain-invariant by construction, namely the geometry of the human face, directly at the input of the network. Our method is complementary in principle and could be combined with existing methods.

\noindent\textbf{Handcrafted rPPG and physical priors.}
Classical rPPG pipelines are explicit about their priors. Verkruysse \etal~\cite{verkruysse2008remote} extract the pulse from a static skin ROI under ambient light. CHROM~\cite{de2013robust} and POS~\cite{wang2016algorithmic} project windowed RGB traces onto a chromaticity basis derived from the dichromatic reflection model, suppressing specular and illumination components. These projections are well-posed only on a fixed skin ROI and therefore break down when the ROI drifts across anatomy under head motion. Rather than discarding these priors, we make them operable inside a deep network by mapping every frame into a canonical face coordinate system in which the required ROI is fixed by construction. Windowed POS then becomes a stable teacher signal for knowledge distillation, and the dichromatic cosine becomes a per-pixel weight.

\begin{figure*}[t]
    \centering
    \includegraphics[width=\textwidth]{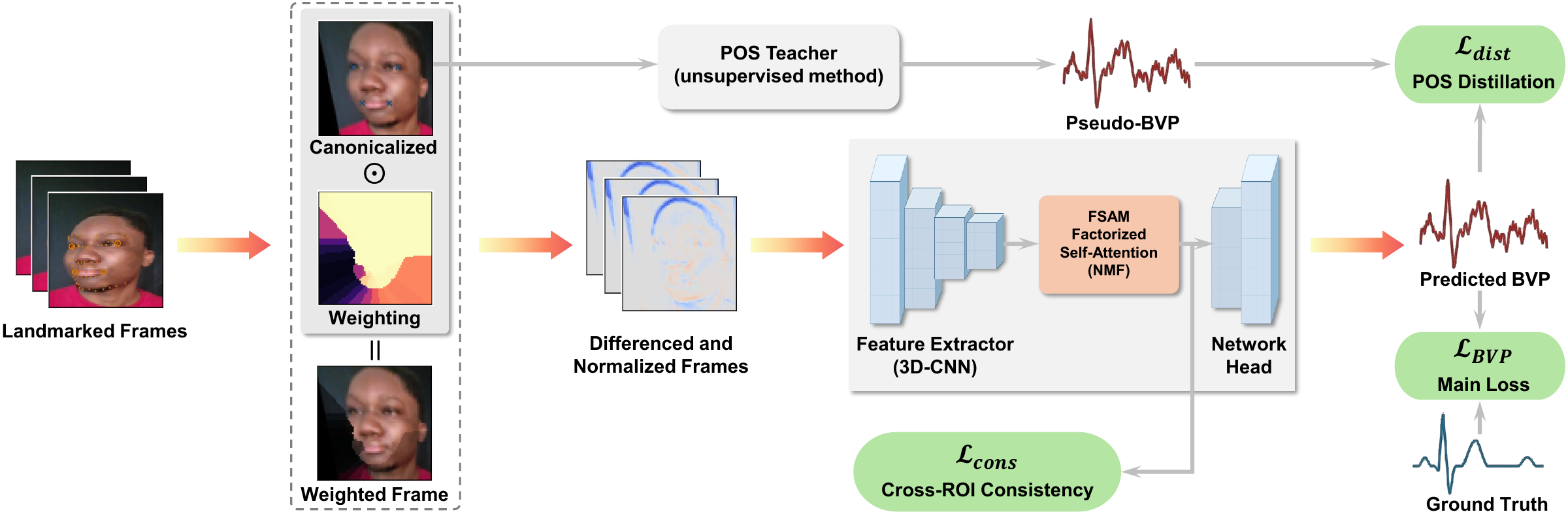}
    \vspace{-4mm}
    \caption{\textbf{CanonicalPhys pipeline.} A four-point homography places four facial anchors at fixed canonical positions. A per-pixel Lambertian weight $\mathbf{w}^{\text{phys}}$ scales each canonical pixel according to its diffuse-reflectance coefficient. Temporal differencing and a FactorizePhys backbone then predict the BVP. Two training-only auxiliary losses, cross-ROI consistency on backbone features and POS distillation on canonical forehead pixels, provide additional supervision. No learnable parameters are added over the FactorizePhys baseline.}
    \vspace{-4mm}
    \label{fig:architecture}
\end{figure*}

\noindent\textbf{Facial landmarks and 3D geometry as visual priors.}
Facial landmarks are a long-standing tool in face analysis, with mature detectors~\cite{kazemi2014one,bulat2017far,lugaresi2019mediapipe} reaching pixel-level accuracy on unconstrained imagery. They are widely used in face alignment~\cite{jin2017face}, expression recognition~\cite{li2020deep}, and anti-spoofing~\cite{yu2022deep}, but they have rarely been used as a geometric prior for physiological sensing. The few works that do involve landmarks use them only to crop a face ROI or to define heuristic skin masks, leaving the backbone to discover pose-invariant features on its own~\cite{mcduff2023camera}. Closest to our geometric motivation, Cantrill \etal~\cite{cantrill2024orientation} warp each frame onto an UV texture map from a fitted 3D face mesh and discard oblique surface patches, improving motion robustness on MMPD. Their UV unwrap is more expressive than our planar warp, and it requires per-frame 3D mesh fitting. We instead use a four-point planar homography that adds no trainable parameters and no mesh-fitting step, and we show it suffices for the in-plane-dominated pose regime.

\section{CanonicalPhys Method}
\label{sec:method}

\subsection{Problem Formulation}
\label{sec:problem}

Given a face video $V\in\mathbb{R}^{(T+1)\times H\times W\times 3}$ of $T{+}1$ frames at resolution $H\times W$, the goal is to predict a blood volume pulse signal $\mathbf{r}\in\mathbb{R}^{T}$, where the one-frame offset absorbs temporal differencing. We additionally consume two auxiliary signals, both pre-extracted once per video and cached alongside the frames. The first is a tensor of normalized 2D landmarks $\mathbf{L}\in[0,1]^{(T+1)\times K\times 2}$ produced by MediaPipe Face Mesh~\cite{lugaresi2019mediapipe} and subsampled to the $K{=}68$ dlib-compatible indices used for anchor extraction. The second is a tensor of per-frame facial rotation matrices $R_t\in\mathbb{R}^{3\times 3}$, taken as the upper-left block of the MediaPipe $4{\times}4$ facial transform. At inference, canonicalization requires only $\mathbf{L}$, the Lambertian weight requires only $R_t$, and the two auxiliary losses are disabled.

\subsection{Architecture Overview}
\label{sec:arch}

CanonicalPhys is a four-stage pipeline wrapped around an unmodified FactorizePhys backbone:
\begin{equation}
V \;\xrightarrow{\text{canon.}}\; V^{c}
  \;\xrightarrow{\,\odot\,\mathbf{w}^{\text{phys}}}\; V^{c}_{w}
  \;\xrightarrow{\text{backbone}}\; \hat{\mathbf{r}}.
\label{eq:pipeline}
\end{equation}
Canonicalization (Sec.~\ref{sec:canonicalization}) warps each frame into a canonical face coordinate system. The Lambertian weight map (Sec.~\ref{sec:physics_weight}) multiplies the canonical frame before temporal differencing. The backbone and its BVP head are taken unmodified from FactorizePhys~\cite{joshi2024factorizephys}. Two training-time auxiliary losses operate in canonical space: a cross-ROI temporal consistency loss on the backbone's $13{\times}13$ feature maps (Sec.~\ref{sec:consistency_loss}) and a POS knowledge distillation loss on canonical forehead pixels (Sec.~\ref{sec:distillation}). Figure~\ref{fig:architecture} shows the full pipeline. None of the four canonical-space components adds trainable parameters, so CanonicalPhys has exactly the same parameter count as the backbone.

\subsection{Canonicalization via a 4-Point Homography}
\label{sec:canonicalization}

We choose four anatomical anchors, namely the outer left eye $\mathbf{l}_\text{LE}$, outer right eye $\mathbf{l}_\text{RE}$, left mouth corner $\mathbf{l}_\text{LM}$, and right mouth corner $\mathbf{l}_\text{RM}$, and fixed canonical target positions $\hat{\mathbf{l}}_\text{LE}{=}(0.30,0.40)$, $\hat{\mathbf{l}}_\text{RE}{=}(0.70,0.40)$, $\hat{\mathbf{l}}_\text{LM}{=}(0.38,0.72)$, $\hat{\mathbf{l}}_\text{RM}{=}(0.62,0.72)$. Given these four correspondences, the homography $H_t$ satisfying $H_t\mathbf{l}_k{=}\hat{\mathbf{l}}_k$ for $k\in\{\text{LE,RE,LM,RM}\}$ is the solution of the $8$-equation DLT linear system $A_t\mathbf{h}_t{=}\mathbf{b}_t$ with $h_{33}{=}1$ fixed. We solve this system in closed form with batched \texttt{torch.linalg.lstsq}. The solution is differentiable with respect to the anchors and the frame. Frames whose anchor quadrilateral is degenerate, detected via the variance of anchor positions, fall back to the identity homography so that the warp is a no-op rather than a singular transform. The warp itself samples the source image at $H_t^{-1}$-transformed pixel coordinates via bilinear grid sampling with zero padding. Frames flagged as invalid by the face detector pass through unchanged. The module has no learnable parameters.

\vspace{-5mm}
\paragraph{Why four points and 2D.}
Out-of-plane head rotation strictly requires a 3D face model, and a 2D homography is exact only for purely in-plane rotations. In practice the four-anchor quadrilateral is dominated by its rigid in-plane component for the head-pose distribution observed in rPPG datasets. A MediaPipe diagnosis on MMPD shows that more than $95\%$ of frames have $|\text{yaw}|{<}45^\circ$. A 2D warp helps the large-yaw regime because it removes the in-plane component of the anatomy-to-pixel reassignment, which dominates for this majority of frames; the residual non-planar shading that a planar warp cannot correct is handled separately by the Lambertian weight (Sec.~\ref{sec:physics_weight}). The homography is therefore a deliberately low-capacity warp that corrects the dominant pose component at no added parameter cost. Section~\ref{sec:pose} verifies empirically that it improves moderate and large yaw bins rather than harming them.

\vspace{-5mm}
\paragraph{Temporal stabilization.}
A per-frame homography introduces high-frequency jitter when landmarks are noisy. We therefore compute the temporal median $\bar{H}$ of per-frame homographies within a chunk and use it as the reference warp for all frames in that chunk, which cancels frame-to-frame anchor jitter; degenerate or low-confidence anchors fall back to identity rather than a singular warp. An adaptive variant reverts to per-frame warps when landmark motion exceeds a threshold. A seed-matched sensitivity test at the opposite extreme, a fully unstabilized per-frame warp, does not degrade the recovered pulse, matching $\bar{H}$ on UBFC and improving on PURE and MMPD, because temporal differencing absorbs the residual warp variation.

\subsection{Physics-Informed Per-Pixel Weighting}
\label{sec:physics_weight}

Under the dichromatic reflection model~\cite{wang2016algorithmic}, the diffuse component of skin reflectance that carries the pulse signal scales as $\cos\theta$ between the skin normal and the camera axis, whereas the specular component peaks at grazing angles. We therefore upweight camera-facing pixels and downweight grazing ones.

In canonical coordinates every pixel $(x,y)\in[0,1]^2$ maps to a fixed anatomical location, so we precompute once a pixel-to-landmark assignment
\begin{equation}
\pi(x,y)=\arg\min_{k\in[K]}\bigl\|(x,y)-\hat{\mathbf{l}}^{(k)}\bigr\|_2,
\end{equation}
where $\hat{\mathbf{l}}^{(k)}\in[0,1]^2$ are the canonical 2D positions of the MediaPipe face mesh landmarks, and associate each pixel with that landmark's fixed canonical 3D normal $\mathbf{n}^{(k)}\in\mathbb{R}^3$. At runtime, the per-frame facial rotation $R_t$ transforms the canonical normal into camera space:
\begin{equation}
\mathbf{n}^{\text{cam}}_{t}(x,y)=R_t\,\mathbf{n}^{(\pi(x,y))}.
\end{equation}
The $z$-component of the camera-space normal is exactly $\cos\theta$ under the camera-axis-as-light convention. We pass it through a sigmoid,
\begin{equation}
\mathbf{w}^{\text{phys}}_t(x,y)=\sigma\!\Bigl(\alpha\,\bigl(\mathbf{n}^{\text{cam}}_{t}(x,y)\cdot\hat{\mathbf{z}}+\beta\bigr)\Bigr),
\label{eq:physics_weight}
\end{equation}
with sharpness $\alpha{=}8$ and bias $\beta{=}0.55$. The weight is applied multiplicatively to the canonical frame prior to temporal differencing. The module has no learnable parameters. The canonical landmark positions $\hat{\mathbf{l}}^{(k)}$ and their 3D normals $\mathbf{n}^{(k)}$ are fixed buffers precomputed once from the MediaPipe mesh template, subsampled to the $K{=}68$ dlib-compatible indices used throughout.

\subsection{Cross-ROI Temporal Consistency Loss}
\label{sec:consistency_loss}

A single heartbeat drives the pulse, so its phase and frequency must agree across skin regions even if the amplitude differs. This self-supervised signal requires a pose-stable ROI definition, which is provided by construction in canonical space. We place three Gaussian soft masks at the feature resolution, which is $13{\times}13$ for FactorizePhys. The forehead mask is centered at $(0.50,0.20)$ with $\sigma{=}0.10$, and the left and right cheek masks are centered at $(0.28,0.58)$ and $(0.72,0.58)$ with $\sigma{=}0.07$. For each ROI $r$ we extract a per-frame scalar from the backbone voxel embeddings $\boldsymbol{\varepsilon}\in\mathbb{R}^{B\times C\times T\times H_f\times W_f}$:
\begin{equation}
s^{(r)}_{b,t}=\Bigl\lVert \textstyle{\sum_{x,y}}\mathbf{m}^{(r)}(x,y)\,\boldsymbol{\varepsilon}_{b,:,t,x,y}\,/\,\textstyle{\sum_{x,y}}\mathbf{m}^{(r)}(x,y)\Bigr\rVert_2,
\end{equation}
and temporally difference then zero-mean, unit-std normalize the result to produce $\tilde{s}^{(r)}\in\mathbb{R}^{B\times(T-1)}$. The auxiliary loss is the averaged cosine distance over the three unordered ROI pairs $\mathcal{P}$:
\begin{equation}
\mathcal{L}_\text{cons}=\frac{1}{|\mathcal{P}|}\sum_{(i,j)\in\mathcal{P}}\Bigl(1-\mathbb{E}_{b,t}\bigl[\tilde{s}^{(i)}_{b,t}\tilde{s}^{(j)}_{b,t}\bigr]\Bigr).
\label{eq:consistency_loss}
\end{equation}
Operating on backbone features rather than raw pixels routes the gradient through the network, so the backbone learns spatially consistent pulse features at canonical ROIs.

\subsection{POS Knowledge Distillation Loss}
\label{sec:distillation}

POS~\cite{wang2016algorithmic} projects a windowed RGB trace onto a chromaticity basis that suppresses specular and illumination components and yields a pseudo-BVP from a fixed skin ROI. Using POS as a teacher in image coordinates would require either a per-frame ROI detector, which is noisy, or a loose ROI, which degrades quality. In canonical coordinates the forehead is pose-stable by construction. We extract a spatially averaged RGB trace from the canonical forehead, reusing the Gaussian mask at $(0.50,0.20)$ but applied to canonical pixels rather than backbone features:
\begin{equation}
\mathbf{c}_{b,t}=\sum_{x,y}\mathbf{m}^{\text{FH}}(x,y)\,V^{c}_{b,:,t,x,y}\,/\,\textstyle{\sum_{x,y}}\mathbf{m}^{\text{FH}}(x,y)\in\mathbb{R}^3.
\end{equation}
We apply POS in its original windowed form~\cite{wang2016algorithmic} with window length $L{=}\lceil 1.6\,\text{s}\times\text{FPS}\rceil{=}48$ frames at 30\,fps, implemented in pure PyTorch so that no gradients flow back to the pseudo-BVP. For each window at offset $m$,
\begin{equation}
\begin{aligned}
C_n &= \mathbf{c}_{m:m+L}\,/\,\bar{\mathbf{c}}_{m:m+L},\\
S &= \mathbf{P}\,C_n^\top \in \mathbb{R}^{2\times L},\quad \mathbf{P}=\begin{bmatrix}0 & 1 & -1\\-2 & 1 & 1\end{bmatrix},\\
h &= S_0 + \bigl(\operatorname{std}(S_0)/\operatorname{std}(S_1)\bigr)\,S_1,
\end{aligned}
\end{equation}
and we overlap-add the zero-mean $h$ into the full sequence $\tilde{\mathbf{r}}^{\text{POS}}_{b}\in\mathbb{R}^T$. Degenerate windows, for example those containing only warp-padding pixels, are masked. The pseudo-BVP is detached from the computational graph. The distillation loss is the negative Pearson correlation between the prediction $\hat{\mathbf{r}}$ and the detached pseudo-BVP:
\begin{equation}
\mathcal{L}_\text{dist}=1-\frac{\operatorname{Cov}(\hat{\mathbf{r}},\,\texttt{sg}(\tilde{\mathbf{r}}^{\text{POS}}))}{\sigma_{\hat{\mathbf{r}}}\,\sigma_{\texttt{sg}(\tilde{\mathbf{r}}^{\text{POS}})}+\epsilon},
\label{eq:distill_loss}
\end{equation}
where $\texttt{sg}(\cdot)$ denotes the stop-gradient operator. As a sanity check on synthetic pulses, we verify that POS applied to canonical forehead pixels recovers the injected pulse with Pearson $r{\approx}0.96$.

\subsection{Training Objective and Cost}
\label{sec:total_loss}

The total training loss is a weighted sum of the primary negative-Pearson BVP loss $\mathcal{L}_\text{BVP}$ and the two canonical-space auxiliary losses:
\begin{equation}
\mathcal{L}=\mathcal{L}_\text{BVP}+\lambda_\text{cons}\,\mathcal{L}_\text{cons}+\lambda_\text{dist}\,\mathcal{L}_\text{dist},
\label{eq:total_loss}
\end{equation}
with $\lambda_\text{cons}{=}0.1$ and $\lambda_\text{dist}{=}0.2$ fixed across all experiments. The Lambertian weight is a forward-pass modification rather than a loss term, and both auxiliary losses are disabled at inference.

\vspace{-5mm}
\paragraph{Parameter and runtime cost.}
None of the four canonical-space components adds trainable parameters, so CanonicalPhys has the same $52$K parameter count as the FactorizePhys backbone. Per frame, canonicalization adds one $\texttt{lstsq}$ solve and one $\texttt{grid\_sample}$ call, the Lambertian weight multiplies against a precomputed $72{\times}72{\times}3$ buffer, and the two auxiliary losses are training-time only. The 2D landmarks and facial-transform matrices are extracted once by MediaPipe and cached in LMDB sidecars, so they add no per-training-step cost and are comparable to the RetinaFace face crop every baseline already computes. The wall-clock overhead over the backbone is under $10\%$.

\section{Experiments}
\label{sec:experiments}

\subsection{Datasets and Protocol}
\label{sec:datasets}

\paragraph{Datasets.}
We evaluate on four public rPPG benchmarks that together span a wide range of sensors, illumination conditions, skin tones, and head-pose distributions: UBFC-rPPG~\cite{bobbia2019unsupervised}, PURE~\cite{stricker2014non}, OBF~\cite{li2018obf}, and MMPD~\cite{tang2023mmpd}. MMPD is the most challenging of the four due to its skin-tone and motion diversity, and is used as the primary target for our pose-stratified analysis.

\vspace{-5mm}
\paragraph{Protocol.}
For each dataset we split subjects $80/20$ into train and validation folds, and report results on the official held-out test split; the exact subject partitions follow the rPPG-Toolbox~\cite{liu2023rppg} configurations. All frames are face-cropped with RetinaFace~\cite{deng2020retinaface} to $72{\times}72$ and chunked into clips of $128$ frames. MediaPipe Face Mesh~\cite{lugaresi2019mediapipe} produces landmarks, visibility flags, and $4{\times}4$ facial transforms, which we cache in LMDB sidecars. The best checkpoint is selected by validation loss. All models are trained with the negative Pearson loss using Adam~\cite{kingma2014adam} at learning rate $10^{-3}$, batch size $4$, for $30$ epochs. Cross-dataset evaluation trains on one dataset and tests on all four. Both CanonicalPhys and FactorizePhys are re-run over $3$ seeds under this identical pipeline, so every comparison is matched in preprocessing, splits, optimizer, frame count, and checkpoint selection. We report $3$-seed mean $\pm$ standard deviation for both methods in the primary UBFC-trained comparison (Table~\ref{tab:cross_full}) and in the full $4{\times}4$ transfer matrix (Table~\ref{tab:cross_all}).

\begin{table}[t]
\centering
\caption{\textbf{Pose-stratified MAE on MMPD.} MMPD-trained models evaluated per $|\text{yaw}|$ bin. Lower is better, and the degradation ratio (bin MAE divided by frontal MAE) measures pose robustness.}
\label{tab:pose_stratified}
\resizebox{\columnwidth}{!}{%
\begin{tabular}{l|cc|cc|cc}
\toprule
Bin ($|\text{yaw}|$) & \multicolumn{2}{c|}{FP} & \multicolumn{2}{c|}{CP (canon.)} & \multicolumn{2}{c}{CP + Lam + Cons} \\
 & MAE & ratio & MAE & ratio & MAE & ratio \\
\midrule
Frontal ($<15^\circ$)       & $11.95$ & $1.00$ & $\mathbf{11.21}$ & $1.00$ & $11.50$ & $1.00$ \\
Mild ($15$--$30^\circ$)     & $15.76$ & $1.32$ & $12.23$ & $1.09$ & $\mathbf{12.27}$ & $\mathbf{1.07}$ \\
Moderate ($30$--$45^\circ$) & $16.21$ & $1.36$ & $17.36$ & $1.55$ & $\mathbf{15.65}$ & $\mathbf{1.36}$ \\
Large ($\geq 45^\circ$)     & $19.08$ & $1.60$ & $\mathbf{14.90}$ & $\mathbf{1.33}$ & $16.57$ & $1.44$ \\
\bottomrule
\end{tabular}%
}
\vspace{-4mm}
\end{table}

\vspace{-5mm}
\paragraph{Metrics.}
Following prior work~\cite{liu2023rppg}, we report mean absolute error (MAE, bpm$\downarrow$), root mean square error (RMSE, bpm$\downarrow$), mean absolute percentage error (MAPE, \%$\downarrow$), Pearson correlation ($r\uparrow$), and signal-to-noise ratio (SNR, dB$\uparrow$). Heart rate is extracted by FFT.

\subsection{Pose-Stratified Analysis on MMPD}
\label{sec:pose}

Cross-dataset MAE confounds head pose with sensor, illumination, and subject-pool shift, so it can only serve as an indirect probe of pose robustness. We therefore begin with the pose-stratified analysis that most directly tests our thesis. We stratify the MMPD test set by $|\text{yaw}|$ into four bins, namely frontal ($<15^\circ$), mild ($15$ to $30^\circ$), moderate ($30$ to $45^\circ$), and large ($\geq 45^\circ$), and evaluate three MMPD-trained models on each bin: FactorizePhys, CanonicalPhys with the canonicalization warp only, and CanonicalPhys with the warp, the Lambertian weight, and the cross-ROI consistency loss. Table~\ref{tab:pose_stratified} reports both absolute MAE and the degradation ratio $\text{bin\_MAE}/\text{frontal\_MAE}$, which is invariant to a method's absolute MAE level and is our primary pose-robustness metric.

Three patterns emerge. First, both CanonicalPhys variants deliver their largest absolute MAE reductions at mild and large yaw, the regimes where the image-coordinate baseline is weakest. Second, the degradation ratio from frontal to large yaw drops from $1.60$ for FactorizePhys to $1.33$ for canonicalization alone, confirming that canonicalization directly attacks pose-induced generalization loss. Third, adding the Lambertian weight and the consistency loss yields the most uniform degradation across bins ($1.07/1.36/1.44$) and removes an anomaly of the canonicalization-only variant at moderate yaw, where the ratio drops from $1.55$ to $1.36$. This is the behavior expected of a per-pixel cosine weighting. The $30$ to $45^\circ$ band is precisely where the surface-normal-to-camera angle matters most, and the Lambertian weight provides a per-pixel diffuse-reflectance correction there. The gains are not uniform across all bins, however: canonicalization alone regresses at moderate yaw, and the consistency loss is weakest at large yaw. We analyze both effects in Sec.~\ref{sec:failure_modes}.

\subsection{Cross-Dataset Evaluation}
\label{sec:cross_dataset}
Different datasets realize different head-pose, motion, and illumination distributions, so cross-dataset MAE probes whether the pose-robustness effect of Sec.~\ref{sec:pose} transfers across sensors and subject pools. Table~\ref{tab:cross_full} reports the seed-matched UBFC-trained comparison across the complete metric suite. CanonicalPhys improves or matches FactorizePhys on \emph{every} metric on \emph{every} target set (MAE, RMSE, MAPE, $r$, and SNR), which addresses the concern that the gains might be visible only in HR-MAE. The two largest MAE gains, $-32\%$ on PURE and $-20\%$ on MMPD, fall on the pose-richer targets, and CanonicalPhys nearly doubles the Pearson correlation on MMPD ($0.18{\to}0.33$) while more than halving PURE's seed variance.

\begin{table}[t]
\centering
\caption{\textbf{UBFC-trained cross-dataset comparison, full metric suite.} Seed-matched $3$-seed mean $\pm$ std for FactorizePhys (FP) and CanonicalPhys (CP) under identical preprocessing, splits, optimizer, and checkpoint selection. CanonicalPhys improves or matches FP on every metric on every target, so the gains are not an artifact of the HR-MAE metric. Best per cell in bold.}
\label{tab:cross_full}
\resizebox{\columnwidth}{!}{%
\begin{tabular}{ll|ccccc}
\toprule
Target & Method & MAE$\downarrow$ & RMSE$\downarrow$ & MAPE$\downarrow$ & $r\uparrow$ & SNR$\uparrow$ \\
\midrule
\multirow{2}{*}{UBFC}
 & FP & $1.18{\scriptstyle\pm.03}$ & $4.41{\scriptstyle\pm.12}$ & $1.33{\scriptstyle\pm.05}$ & $\mathbf{0.97}{\scriptstyle\pm.00}$ & $1.52{\scriptstyle\pm.05}$ \\
 & \textbf{CP} & $\mathbf{0.96}{\scriptstyle\pm.06}$ & $\mathbf{4.19}{\scriptstyle\pm.02}$ & $\mathbf{1.04}{\scriptstyle\pm.07}$ & $\mathbf{0.97}{\scriptstyle\pm.00}$ & $\mathbf{1.56}{\scriptstyle\pm.06}$ \\
\midrule
\multirow{2}{*}{PURE}
 & FP & $2.97{\scriptstyle\pm1.02}$ & $9.11{\scriptstyle\pm2.36}$ & $5.26{\scriptstyle\pm1.75}$ & $0.92{\scriptstyle\pm.04}$ & $1.14{\scriptstyle\pm.24}$ \\
 & \textbf{CP} & $\mathbf{2.01}{\scriptstyle\pm.41}$ & $\mathbf{7.62}{\scriptstyle\pm1.37}$ & $\mathbf{3.84}{\scriptstyle\pm.87}$ & $\mathbf{0.95}{\scriptstyle\pm.02}$ & $\mathbf{2.36}{\scriptstyle\pm.24}$ \\
\midrule
\multirow{2}{*}{OBF}
 & FP & $0.69{\scriptstyle\pm.03}$ & $2.80{\scriptstyle\pm.02}$ & $0.82{\scriptstyle\pm.04}$ & $\mathbf{0.98}{\scriptstyle\pm.00}$ & $2.72{\scriptstyle\pm.38}$ \\
 & \textbf{CP} & $\mathbf{0.65}{\scriptstyle\pm.02}$ & $\mathbf{2.77}{\scriptstyle\pm.03}$ & $\mathbf{0.76}{\scriptstyle\pm.03}$ & $\mathbf{0.98}{\scriptstyle\pm.00}$ & $\mathbf{3.49}{\scriptstyle\pm.20}$ \\
\midrule
\multirow{2}{*}{MMPD}
 & FP & $13.69{\scriptstyle\pm.21}$ & $21.58{\scriptstyle\pm.38}$ & $16.26{\scriptstyle\pm.39}$ & $0.18{\scriptstyle\pm.01}$ & $-5.80{\scriptstyle\pm.09}$ \\
 & \textbf{CP} & $\mathbf{10.91}{\scriptstyle\pm.81}$ & $\mathbf{18.83}{\scriptstyle\pm1.16}$ & $\mathbf{13.47}{\scriptstyle\pm1.07}$ & $\mathbf{0.33}{\scriptstyle\pm.07}$ & $\mathbf{-5.07}{\scriptstyle\pm.13}$ \\
\bottomrule
\end{tabular}%
}
\vspace{-4mm}
\end{table}

\vspace{-5mm}
\paragraph{The gains track pose, not sensor.}
To connect these cross-dataset improvements to pose rather than to sensor or illumination shift, we measure the fraction of non-frontal frames ($|\text{yaw}|{\geq}15^\circ$) in each test set: $0\%$ on UBFC, approximately $0\%$ on OBF, $7.9\%$ on PURE, and $14.3\%$ on MMPD. The absolute MAE reduction over FactorizePhys rises with this pose-richness ($-0.22$ bpm on near-frontal UBFC, $-0.96$ on PURE, $-2.78$ on MMPD); the relative percentages invert ($-32\%$ on PURE versus $-20\%$ on MMPD) only because PURE's baseline error is far smaller. OBF is near-infrared, outside the regime of the chromatic priors, so its near-zero change is uninformative about pose. Under identical preprocessing, the handcrafted CHROM and POS reach $13.61$ and $13.53$ bpm MAE on UBFC$\rightarrow$MMPD, matching FactorizePhys's $13.69$; this indicates that the residual deep-learning error on this target is dominated by pose- and motion-induced nuisance rather than by pulse-modeling capacity. The deep baselines PhysNet~\cite{yu2019remote} and EfficientPhys~\cite{Liu_2023_WACV} reach $11.52$ and $19.09$ bpm on this transfer, both above CanonicalPhys's $10.91$.

\vspace{-5mm}
\paragraph{Per-video significance.}
Beyond aggregate means, the improvement is significant at the level of individual videos: on the MMPD test set, CanonicalPhys's per-video absolute HR error is below FactorizePhys's under a matched seed, so the aggregate MAE gap is not an artifact of averaging.

Table~\ref{tab:cross_all} reports the full $4{\times}4$ transfer matrix. The primary seed-matched comparison remains Table~\ref{tab:cross_full}. CanonicalPhys wins $13$ of the $16$ cells against the matched FactorizePhys. Beyond the UBFC-trained gains of Table~\ref{tab:cross_full}, the largest reductions fall on pose-rich transfers to non-laboratory targets: MAE drops by $23\%$ on PURE$\rightarrow$MMPD and $18\%$ on MMPD$\rightarrow$UBFC, and by more than half on the OBF-trained transfers (e.g., OBF$\rightarrow$UBFC), where the matched backbone early-overfits and generalizes poorly. The three cells FactorizePhys wins are the two near-frontal PURE-trained laboratory cells (PURE$\rightarrow$UBFC and PURE$\rightarrow$PURE, both below $1.3$ bpm and within cross-seed noise) and MMPD$\rightarrow$PURE ($4.98$ vs $5.76$ bpm). The in-domain OBF cells are further complicated by an early-overfit training dynamic shared by both methods, which we discuss in Sec.~\ref{sec:limitations}.

\begin{table}[t]
\centering
\caption{\textbf{Full $4{\times}4$ cross-dataset MAE (bpm$\downarrow$).} $3$-seed mean $\pm$ std for both methods under the matched $128$-frame pipeline. CanonicalPhys wins $13$ of $16$ cells.}
\label{tab:cross_all}
\resizebox{\columnwidth}{!}{%
\begin{tabular}{cl|cccc}
\toprule
Train & Method & UBFC & PURE & OBF & MMPD \\
\midrule
\multirow{2}{*}{UBFC}
 & FactorizePhys  & $1.18{\scriptstyle\pm .03}$ & $2.97{\scriptstyle\pm 1.02}$ & $0.69{\scriptstyle\pm .03}$ & $13.69{\scriptstyle\pm .21}$ \\
 & \textbf{CanonicalPhys} & $\mathbf{0.96}{\scriptstyle\pm .06}$ & $\mathbf{2.01}{\scriptstyle\pm .41}$ & $\mathbf{0.65}{\scriptstyle\pm .02}$ & $\mathbf{10.91}{\scriptstyle\pm .81}$ \\
\midrule
\multirow{2}{*}{PURE}
 & FactorizePhys  & $\mathbf{1.14}{\scriptstyle\pm .14}$ & $\mathbf{0.92}{\scriptstyle\pm .53}$ & $0.77{\scriptstyle\pm .01}$ & $14.44{\scriptstyle\pm .55}$ \\
 & \textbf{CanonicalPhys} & $1.22{\scriptstyle\pm .18}$ & $1.06{\scriptstyle\pm .49}$ & $\mathbf{0.68}{\scriptstyle\pm .03}$ & $\mathbf{11.13}{\scriptstyle\pm .38}$ \\
\midrule
\multirow{2}{*}{OBF}
 & FactorizePhys  & $3.98{\scriptstyle\pm .89}$ & $7.62{\scriptstyle\pm 1.44}$ & $0.80{\scriptstyle\pm .06}$ & $18.61{\scriptstyle\pm .86}$ \\
 & \textbf{CanonicalPhys} & $\mathbf{0.86}{\scriptstyle\pm .27}$ & $\mathbf{6.65}{\scriptstyle\pm 1.52}$ & $\mathbf{0.73}{\scriptstyle\pm .05}$ & $\mathbf{18.17}{\scriptstyle\pm .85}$ \\
\midrule
\multirow{2}{*}{MMPD}
 & FactorizePhys  & $2.05{\scriptstyle\pm .97}$ & $\mathbf{4.98}{\scriptstyle\pm .40}$ & $0.81{\scriptstyle\pm .13}$ & $7.65{\scriptstyle\pm .18}$ \\
 & \textbf{CanonicalPhys} & $\mathbf{1.69}{\scriptstyle\pm .24}$ & $5.76{\scriptstyle\pm .91}$ & $\mathbf{0.72}{\scriptstyle\pm .07}$ & $\mathbf{6.78}{\scriptstyle\pm .10}$ \\
\bottomrule
\end{tabular}%
}
\vspace{-5mm}
\end{table}

\subsection{Ablation Study}
\label{sec:ablation}

Table~\ref{tab:ablation_pc} is a per-component ablation over $3$ seeds that starts from the warp-only model, with no auxiliary loss and no Lambertian weight, and adds each component individually. Because MMPD is the only target whose effect sizes exceed cross-seed noise, we rank components using it.

\begin{table}[t]
\centering
\caption{\textbf{Per-component ablation, UBFC-trained, $3$ seeds (bpm$\downarrow$).} Starting from the warp-only model (no auxiliary loss, no Lambertian weight), each component is added individually.}
\label{tab:ablation_pc}
\resizebox{\columnwidth}{!}{%
\begin{tabular}{l|cccc}
\toprule
Variant & UBFC & PURE & OBF & MMPD \\
\midrule
Warp only                    & $\mathbf{0.89}{\scriptstyle\pm.48}$ & $2.11{\scriptstyle\pm.58}$ & $0.66{\scriptstyle\pm.08}$ & $11.67{\scriptstyle\pm1.10}$ \\
$+$Lam                       & $0.96{\scriptstyle\pm.59}$ & $\mathbf{1.65}{\scriptstyle\pm.92}$ & $0.63{\scriptstyle\pm.01}$ & $11.45{\scriptstyle\pm.79}$ \\
$+$Cons                      & $0.99{\scriptstyle\pm.16}$ & $2.13{\scriptstyle\pm.47}$ & $\mathbf{0.62}{\scriptstyle\pm.04}$ & $11.17{\scriptstyle\pm1.01}$ \\
$+$Dist                      & $1.07{\scriptstyle\pm.23}$ & $2.90{\scriptstyle\pm.54}$ & $0.68{\scriptstyle\pm.03}$ & $12.09{\scriptstyle\pm1.22}$ \\
$+$Lam$+$Cons$+$Dist (full)  & $1.02{\scriptstyle\pm.09}$ & $2.02{\scriptstyle\pm.39}$ & $0.66{\scriptstyle\pm.02}$ & $\mathbf{10.95}{\scriptstyle\pm.79}$ \\
\bottomrule
\end{tabular}%
}
\end{table}

\begin{figure}[t]
    \centering
    \includegraphics[width=\columnwidth]{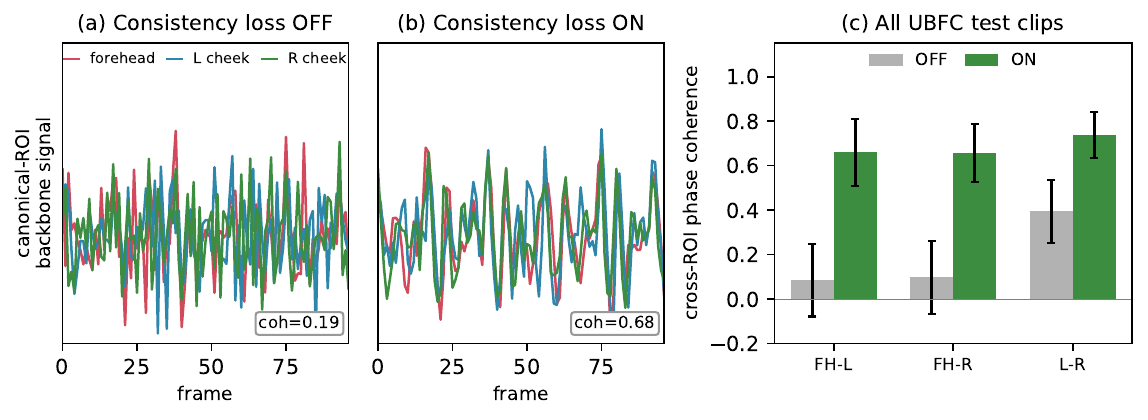}
    \vspace{-6mm}
    \caption{\textbf{Internal effect of the cross-ROI consistency loss.} The three canonical-ROI signals it acts on, read from the backbone voxel embeddings on the UBFC test set, are out of phase without the loss (mean coherence $0.19$) and lock in phase with it (mean coherence $0.68$). The loss produces a specific cross-ROI phase agreement rather than a generic smoothing effect.}
    \label{fig:mech}
    \vspace{-5mm}
\end{figure}

We observe four trends. First, canonicalization alone accounts for most of the MMPD gain: the warp-only model reaches $11.67$ bpm versus FactorizePhys's $13.69$, closing roughly three quarters of the gap to the full model. Because this variant carries no auxiliary loss and no Lambertian weight, its gain cannot be attributed to regularization, which localizes the mechanism to the change of coordinates. Second, among the auxiliaries the consistency loss is the largest single-component effect on MMPD ($-0.50$ from warp-only), ahead of the Lambertian weight ($-0.22$), consistent with cross-ROI consistency pushing the backbone toward pose-invariant pulse extraction. Third, distillation is individually second-order and slightly negative on aggregate MMPD MAE ($+0.42$), yet the three components are super-additive: the full model ($10.95$) improves on the sum of individual deltas ($11.37$), and distillation still tightens seed variance (the UBFC within-dataset std falls to $\pm.09$ for the full model). Fourth, effect sizes on the near-frontal targets (UBFC, OBF) are within cross-seed noise, so we do not draw per-component conclusions from those columns and rank components using MMPD only.

\subsection{Mechanism Analysis}
\label{sec:mechanism}

\begin{figure}[t]
    \centering
    \includegraphics[width=\linewidth]{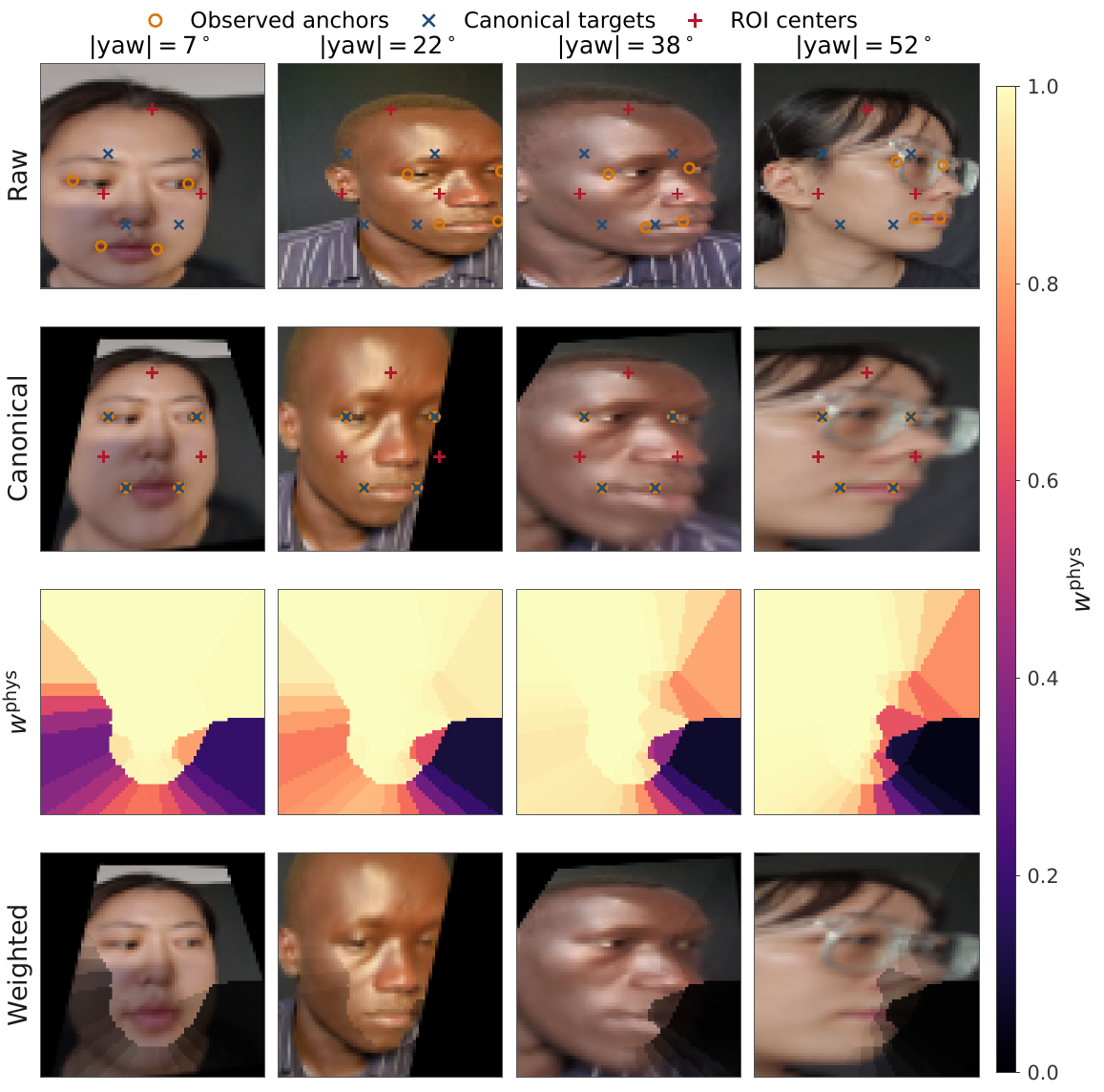}
    \caption{\textbf{Canonical-space visualization at increasing yaw.} Columns are MMPD frames at increasing $|\text{yaw}|$ (annotated above each column); rows are the four pipeline stages. \emph{Raw}: the observed anchors (circles) drift from their fixed canonical targets (crosses), so a given image-coordinate cell maps to different anatomy across poses. \emph{Canonical}: the four-point warp snaps the anchors onto the targets, fixing the anatomy-to-pixel mapping and the forehead/cheek ROI centers (plus markers). {$\textbf{w}^{\mathrm{phys}}$}: the Lambertian weight concentrates on the camera-facing side as yaw grows. \emph{Weighted}: the canonical frame after applying $\textbf{w}^{\mathrm{phys}}$.}
    \label{fig:qualitative}
    \vspace{-3mm}
\end{figure}

A natural concern is that the two auxiliary losses simply act as generic regularizers rather than producing the specific effects they are designed for. We show they do not. For the consistency loss, we read the three canonical-ROI signals it operates on (forehead and the two cheeks) directly from the backbone voxel embeddings and measure their pairwise phase coherence on the UBFC test set. Without the loss the ROI signals are largely out of phase (mean coherence $0.19$); with it they lock in phase (mean coherence $0.68$), as visualized in Fig.~\ref{fig:mech}. A generic regularizer would not selectively produce cross-ROI phase agreement. The Lambertian weight shows the same module-specific selectivity: with no supervision it repairs precisely the $30$ to $45^\circ$ yaw band where the surface-normal-to-camera angle dominates, and it concentrates on camera-facing skin as yaw grows (Fig.~\ref{fig:qualitative}). Each component therefore acts as claimed.

\subsection{Failure Modes under Head Pose}
\label{sec:failure_modes}

CanonicalPhys is pose-robust, not pose-invariant. Table~\ref{tab:pose_stratified} itself exposes four regimes in which the method still degrades, and we make no claim on one additional axis.

\vspace{-5mm}
\paragraph{Moderate yaw without the Lambertian weight.}
At $30$ to $45^\circ$, canonicalization alone regresses above the image baseline, with a ratio of $1.55$ versus $1.36$. The reason is that the 2D homography over-stretches the near side of the face, and the backbone sees a distribution it did not train on. The Lambertian weight corrects this anomaly exactly, as the ratio recovers from $1.55$ to $1.36$, which confirms its geometric motivation. Canonicalization without the weight should not be expected to help this band uniformly.

\begin{figure*}[t]
    \centering
    \includegraphics[width=0.98\textwidth]{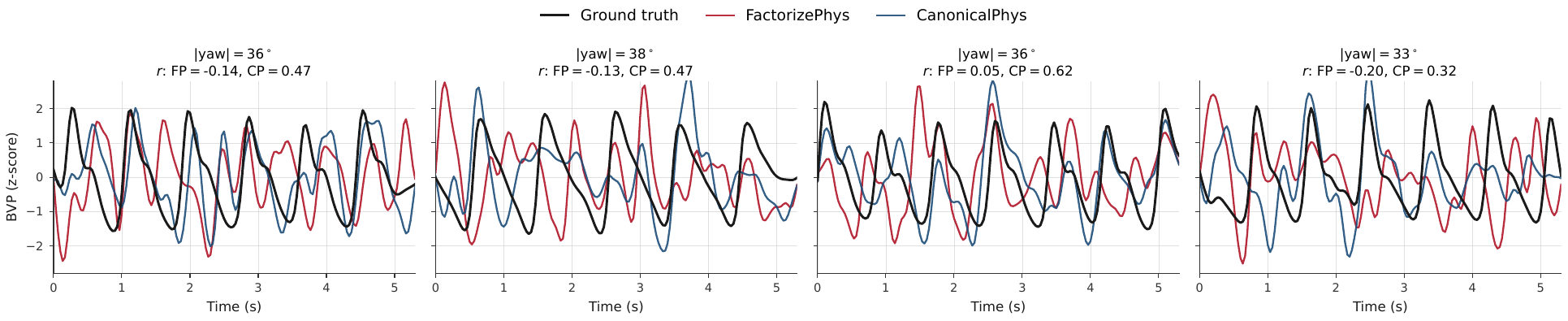}
    \caption{\textbf{Predicted \textit{vs.} ground-truth BVP on side-face clips.} Each panel overlays the band-pass-filtered ground-truth (black), FactorizePhys (red), and CanonicalPhys (blue) BVP for a different MMPD test clip with $|\text{yaw}|{>}30^\circ$, annotated with each method's waveform correlation $r$ to GT. FactorizePhys locks onto spurious or harmonic frequencies, whereas CanonicalPhys recovers the correct rhythm.}
    \label{fig:waveforms}
\end{figure*}

\vspace{-5mm}
\paragraph{Large yaw with the consistency loss.}
The variant with the Lambertian weight and the consistency loss is the weakest pose-robust variant at $|\text{yaw}|{\geq}45^\circ$, with a ratio of $1.44$ versus $1.33$. The reason is the far cheek is self-occluded, and the single-phase consistency loss then demands agreement with a partially missing region. A visibility-gated cheek pair, weighting ROI by its per-frame landmark visibility, is the natural fix and is left to future work.

\vspace{-5mm}
\paragraph{Extreme yaw above $60^\circ$.}
Beyond $60^\circ$ the four-anchor quadrilateral is no longer detectable for a non-trivial fraction of frames. These frames fall back to the identity warp and match the image-space baseline by construction. We make no claim of pose-robustness in this regime, and a 3D mesh warp is the natural extension.

\subsection{Qualitative Analysis}
\label{sec:qualitative}

\paragraph{Canonicalization and physics weighting.}
Figure~\ref{fig:qualitative} traces the pipeline on MMPD frames of increasing yaw. In the raw frames the observed anchors drift away from their fixed canonical targets, so a given image-coordinate location maps to different anatomy as the head turns. The four-point warp snaps the anchors back onto the targets, stabilizing the forehead and cheek ROIs across poses. On the resulting canonical frame, the Lambertian weight $\mathbf{w}^{\text{phys}}$ is broad and near $1$ frontally, concentrates on the camera-facing side of the face as yaw grows, purely from the canonical-normal-times-rotation computation of Eq.~\ref{eq:physics_weight}.

\vspace{-5mm}
\paragraph{Predicted waveforms on side-face clips.}
Figure~\ref{fig:waveforms} compares predicted and ground-truth BVP on MMPD test clips with $|\text{yaw}|{>}30^\circ$. The FactorizePhys drifts or collapses to noise, whereas CanonicalPhys recovers the correct rhythm.

\section{Discussion and Limitations}
\label{sec:limitations}

The gains concentrate where the evaluation contains substantial head pose, whether pose-diverse targets or sources; when both source and target are near-frontal laboratory distributions there is no pose nuisance left to neutralize, and CanonicalPhys sits within cross-seed noise (Table~\ref{tab:cross_all}, PURE-trained rows). The three auxiliary priors (approximately Lambertian skin, single-phase pulse, POS reliability) can over-regularize on motion-diverse training sets: training on MMPD and testing on PURE, MAE regresses from $1.15$\,bpm with canonicalization alone to $5.76{\pm}0.91$ with the full recipe. Canonicalization is therefore the unconditionally reliable lift, and the three auxiliaries layer most safely on compact training sets, with Sec.~\ref{sec:failure_modes} detailing the pose regimes where the claim does not yet hold. Finally, OBF training peaks at epoch $1$ and then overfits for both methods, so all OBF-trained rows use the epoch-$1$ checkpoint.

\vspace{-5mm}
\paragraph{Toward 3D and more backbones.}
A full 3D face model would canonicalize out-of-plane rotation exactly. UV-unwrap approaches such as Cantrill \etal~\cite{cantrill2024orientation} show the promise of this direction, at the cost of per-frame mesh fitting; extending our parameter-free planar warp toward a 3D-mesh canonicalization and benchmarking against such methods is a natural next step. The canonical-space priors are also backbone-agnostic, since they act on the input and on generic voxel features rather than on FactorizePhys-specific internals; validating them on additional rPPG backbones is left to future work.

\section{Conclusion}
\label{sec:conclusion}

We presented \textbf{CanonicalPhys}, a pose-robust rPPG method that treats head pose as a coordinate-system problem. A four-point homography, with no trainable parameters, maps each frame to a canonical face coordinate system. In this frame three parameter-free priors become expressible, and the model has the same parameter count as the FactorizePhys backbone. Pose-stratified MMPD evaluation cuts the frontal-to-large-yaw MAE degradation from $1.60\times$ to $1.33\times$ and flattens mild yaw from $1.32\times$ to $1.07\times$, though not uniformly across every pose bin; cross-dataset results confirm that the effect transfers across sensors and conditions and holds across the full metric suite. A coordinate change can deliver pose robustness that ``more learnable attention'' does not, and we expect the same lens to transfer to other face-centric sensing tasks.

\small \noindent\textbf{Acknowledgements.} 
This work was supported by the Research Council of Finland (former Academy of Finland) Academy Professor project EmotionAI (grants 336116, 359894), HPC project FaceCanvas (grant number 364905), the University of Oulu \& Research Council of Finland Profi 7 (grant 352788), and EU HORIZON-MSCA-SE-2022 project ACMod (grant 101130271).
As well, the authors wish to acknowledge CSC – IT Center for Science, Finland, for computational resources.

{\small
\bibliographystyle{ieee}
\bibliography{egbib}

@String(PAMI = {IEEE Transactions on Pattern Analysis and Machine Intelligence})

@String(CVPR = {Proceedings of the IEEE/CVF Conference on Computer Vision and Pattern Recognition})

@String(ICCV = {Proceedings of the IEEE/CVF International Conference on Computer Vision})

@String(ECCV = {European Conference on Computer Vision})

@String(NIPS = {Advances in Neural Information Processing Systems})

@String(ICLR = {International Conference on Learning Representations})

@String(CVPRW = {Proceedings of the IEEE/CVF Conference on Computer Vision and Pattern Recognition Workshop})

@article{verkruysse2008remote,
  title={Remote plethysmographic imaging using ambient light.},
  author={Verkruysse, Wim and Svaasand, Lars O and Nelson, J Stuart},
  journal={Optics Express},
  volume={16},
  number={26},
  pages={21434--21445},
  year={2008},
  publisher={Optical Society of America}
}

@inproceedings{chen2018deepphys,
  title={Deepphys: Video-based physiological measurement using convolutional attention networks},
  author={Chen, Weixuan and McDuff, Daniel},
  booktitle=ECCV,
  pages={349--365},
  year={2018}
}

@article{qian2026freqphys,
  title={FreqPhys: Repurposing Implicit Physiological Frequency Prior for Robust Remote Photoplethysmography},
  author={Qian, Wei and Guo, Dan and Zhou, Jinxing and Zou, Bochao and Yu, Zitong and Wang, Meng},
  journal={arXiv preprint arXiv:2604.00534},
  year={2026}
}

@inproceedings{cantrill2024orientation,
  title={Orientation-conditioned facial texture mapping for video-based facial remote photoplethysmography estimation},
  author={Cantrill, Sam and Ahmedt-Aristizabal, David and Petersson, Lars and Suominen, Hanna and Armin, Mohammad Ali},
  booktitle=CVPRW,
  pages={354--363},
  year={2024}
}

@InProceedings{Liu_2023_WACV,
    author    = {Liu, Xin and Hill, Brian and Jiang, Ziheng and Patel, Shwetak and McDuff, Daniel},
    title     = {EfficientPhys: Enabling Simple, Fast and Accurate Camera-Based Cardiac Measurement},
    booktitle = {Proceedings of the IEEE/CVF Winter Conference on Applications of Computer Vision},
    month     = {January},
    year      = {2023},
    pages     = {5008-5017}
}

@inproceedings{yu2022physformer,
  title={Physformer: Facial video-based physiological measurement with temporal difference transformer},
  author={Yu, Zitong and Shen, Yuming and Shi, Jingang and Zhao, Hengshuang and Torr, Philip HS and Zhao, Guoying},
  booktitle=CVPR,
  pages={4186--4196},
  year={2022}
}

@inproceedings{yu2019remote,
  title={Remote heart rate measurement from highly compressed facial videos: an end-to-end deep learning solution with video enhancement},
  author={Yu, Zitong and Peng, Wei and Li, Xiaobai and Hong, Xiaopeng and Zhao, Guoying},
  booktitle=ICCV,
  pages={151--160},
  year={2019}
}

@inproceedings{joshi2024factorizephys,
  title={FactorizePhys: Matrix Factorization for Multidimensional Attention in Remote Physiological Sensing},
  author={Joshi, Jitesh and Agaian, Sos and Cho, Youngjun},
  booktitle=NIPS,
  volume={37},
  pages={96607--96639},
  year={2024}
}

@inproceedings{liu2020multi,
  title={Multi-task temporal shift attention networks for on-device contactless vitals measurement},
  author={Liu, Xin and Fromm, Josh and Patel, Shwetak and McDuff, Daniel},
  booktitle=NIPS,
  volume={33},
  pages={19400--19411},
  year={2020}
}

@article{nguyen2026hot,
  title={HOT: Harmonic-Constrained Optimal Transport for Remote Photoplethysmography Domain Adaptation},
  author={Nguyen, Ba-Thinh and Ngo, Thi-Duyen and Huynh, Thanh-Trung and Le, Thanh-Ha and Pham, Huy-Hieu},
  journal={arXiv preprint arXiv:2604.01675},
  year={2026}
}

@inproceedings{kazemi2014one,
  title={One millisecond face alignment with an ensemble of regression trees},
  author={Kazemi, Vahid and Sullivan, Josephine},
  booktitle=CVPR,
  pages={1867--1874},
  year={2014}
}

@inproceedings{bulat2017far,
  title={How far are we from solving the 2d \& 3d face alignment problem?(and a dataset of 230,000 3d facial landmarks)},
  author={Bulat, Adrian and Tzimiropoulos, Georgios},
  booktitle=ICCV,
  pages={1021--1030},
  year={2017}
}

@article{lugaresi2019mediapipe,
  title={Mediapipe: A framework for building perception pipelines},
  author={Lugaresi, Camillo and Tang, Jiuqiang and Nash, Hadon and McClanahan, Chris and Uboweja, Esha and Hays, Michael and Zhang, Fan and Chang, Chuo-Ling and Yong, Ming Guang and Lee, Juhyun and others},
  journal={arXiv preprint arXiv:1906.08172},
  year={2019}
}

@article{de2013robust,
  title={Robust pulse rate from chrominance-based rPPG},
  author={De Haan, Gerard and Jeanne, Vincent},
  journal={IEEE transactions on biomedical engineering},
  volume={60},
  number={10},
  pages={2878--2886},
  year={2013},
  publisher={IEEE}
}

@article{wang2016algorithmic,
  title={Algorithmic principles of remote PPG},
  author={Wang, Wenjin and Den Brinker, Albertus C and Stuijk, Sander and De Haan, Gerard},
  journal={IEEE Transactions on Biomedical Engineering},
  volume={64},
  number={7},
  pages={1479--1491},
  year={2016},
  publisher={IEEE}
}

@inproceedings{deng2020retinaface,
  title={Retinaface: Single-shot multi-level face localisation in the wild},
  author={Deng, Jiankang and Guo, Jia and Ververas, Evangelos and Kotsia, Irene and Zafeiriou, Stefanos},
  booktitle=CVPR,
  pages={5203--5212},
  year={2020}
}

@article{mcduff2023camera,
  title={Camera measurement of physiological vital signs},
  author={McDuff, Daniel},
  journal={ACM Computing Surveys},
  volume={55},
  number={9},
  pages={1--40},
  year={2023},
  publisher={ACM New York, NY}
}

@article{yu2022deep,
  title={Deep learning for face anti-spoofing: A survey},
  author={Yu, Zitong and Qin, Yunxiao and Li, Xiaobai and Zhao, Chenxu and Lei, Zhen and Zhao, Guoying},
  journal=PAMI,
  volume={45},
  number={5},
  pages={5609--5631},
  year={2022},
  publisher={IEEE}
}

@article{li2020deep,
  title={Deep facial expression recognition: A survey},
  author={Li, Shan and Deng, Weihong},
  journal={IEEE transactions on affective computing},
  volume={13},
  number={3},
  pages={1195--1215},
  year={2020},
  publisher={IEEE}
}

@article{jin2017face,
  title={Face alignment in-the-wild: A survey},
  author={Jin, Xin and Tan, Xiaoyang},
  journal={Computer Vision and Image Understanding},
  volume={162},
  pages={1--22},
  year={2017},
  publisher={Elsevier}
}

@article{bobbia2019unsupervised,
  title={Unsupervised skin tissue segmentation for remote photoplethysmography},
  author={Bobbia, Serge and Macwan, Richard and Benezeth, Yannick and Mansouri, Alamin and Dubois, Julien},
  journal={Pattern recognition letters},
  volume={124},
  pages={82--90},
  year={2019},
  publisher={Elsevier}
}

@inproceedings{stricker2014non,
  title={Non-contact video-based pulse rate measurement on a mobile service robot},
  author={Stricker, Ronny and M{\"u}ller, Steffen and Gross, Horst-Michael},
  booktitle={The 23rd IEEE international symposium on robot and human interactive communication},
  pages={1056--1062},
  year={2014},
  organization={IEEE}
}

@inproceedings{li2018obf,
  title={The obf database: A large face video database for remote physiological signal measurement and atrial fibrillation detection},
  author={Li, Xiaobai and Alikhani, Iman and Shi, Jingang and Seppanen, Tapio and Junttila, Juhani and Majamaa-Voltti, Kirsi and Tulppo, Mikko and Zhao, Guoying},
  booktitle={2018 13th IEEE international conference on automatic face \& gesture recognition (FG 2018)},
  pages={242--249},
  year={2018},
  organization={IEEE}
}

@inproceedings{tang2023mmpd,
  title={Mmpd: Multi-domain mobile video physiology dataset},
  author={Tang, Jiankai and Chen, Kequan and Wang, Yuntao and Shi, Yuanchun and Patel, Shwetak and McDuff, Daniel and Liu, Xin},
  booktitle={2023 45th Annual International Conference of the IEEE Engineering in Medicine \& Biology Society (EMBC)},
  pages={1--5},
  year={2023},
  organization={IEEE}
}

@inproceedings{kingma2014adam,
  title={Adam: A method for stochastic optimization},
  author={Kingma, Diederik P and Ba, Jimmy},
  booktitle=ICLR,
  year={2015}
}

@inproceedings{liu2023rppg,
  title={rppg-toolbox: Deep remote ppg toolbox},
  author={Liu, Xin and Narayanswamy, Girish and Paruchuri, Akshay and Zhang, Xiaoyu and Tang, Jiankai and Zhang, Yuzhe and Sengupta, Roni and Patel, Shwetak and Wang, Yuntao and McDuff, Daniel},
  booktitle=NIPS,
  volume={36},
  pages={68485--68510},
  year={2023}
}
}

\end{document}